\pdfoutput=1

\documentclass[11pt]{article}

\usepackage[final]{acl}

\usepackage{ulem}

\usepackage{times}
\usepackage{latexsym}
\usepackage{graphicx}
\usepackage{amsmath}
\usepackage{amssymb}
\usepackage{multirow}
\usepackage[T1]{fontenc}
\usepackage{booktabs} 
\usepackage{tabularx}
\usepackage{makecell}
\usepackage[utf8]{inputenc}

\usepackage{microtype}

\usepackage{inconsolata}

\title{Not All Experts are Equal: Efficient Expert Pruning and Skipping for Mixture-of-Experts Large Language Models}

\author{Xudong Lu$^{*1}$, Qi Liu$^{*2,4}$, Yuhui Xu$^{3}$, Aojun Zhou$^{\ddag 1}$, Siyuan Huang$^{2,4}$\\ {\bf Bo Zhang}$^{4}$, {\bf Junchi Yan}$^{2,4}$, {\bf Hongsheng Li}$^{\dagger 1,5}$\\
 $^1$CUHK MMLab\quad $^2$Shanghai Jiao Tong University\quad  $^3$Salesforce AI Research\\ $^4$Shanghai Artificial Intelligence Laboratory\quad $^5$CPII under InnoHK\\
  \texttt{\{luxudong@link,hsli@ee\}.cuhk.edu.hk}\\
  \texttt{purewhite@sjtu.edu.cn}\quad \texttt{\{xyh6666,aojunzhou\}@gmail.com}
}

\begin{document}
\maketitle
\newcommand\blfootnote[1]{%
\begingroup
\renewcommand\thefootnote{}\footnote{#1}%
\addtocounter{footnote}{-1}%
\endgroup
}
\begin{abstract}
A pivotal advancement in the progress of large language models (LLMs) is the emergence of the Mixture-of-Experts (MoE) LLMs. Compared to traditional LLMs, MoE LLMs can achieve higher performance with fewer active parameters, but it is still hard to deploy them due to their immense parameter sizes. Different from previous weight pruning methods that rely on specifically designed hardware, this paper mainly aims to enhance the deployment efficiency of MoE LLMs by introducing plug-and-play expert-level sparsification techniques. Specifically, we propose, for the first time to our best knowledge, post-training approaches for task-agnostic and task-specific expert pruning and skipping of MoE LLMs, tailored to improve deployment efficiency while maintaining model performance across a wide range of tasks. Extensive experiments show that our proposed methods can simultaneously reduce model sizes and increase the inference speed, while maintaining satisfactory performance. Code will be made available at \url{https://github.com/Lucky-Lance/Expert_Sparsity}.
\end{abstract}

\blfootnote{$^*$Equal contribution\quad $^\dagger$Corresponding author
$^\ddag$Project lead}

\vspace{-2em}
\section{Introduction}\label{sec:intro}

Large language models (LLMs) have shown remarkable abilities across various domains~\cite{OpenAI2023GPT4TR, zhou2024solving}, as evidenced by the widespread use of ChatGPT and Gemini~\cite{team2023gemini}. Recent notable advancement in this area is the introduction of the open-sourced Mixture-of-Experts (MoE) LLM, Mixtral 8x7B~\cite{jiang2024mixtral}, which sparsely activates only a portion of its parameters during the training and inference process. This model surpasses the performance of dense Transformer-based LLMs, such as LLaMA-2 70B~\cite{touvron2023llama1,touvron2023llama2}, with fewer active parameters (13B) during inference. 

MoE LLMs achieve a reduction in on-the-fly (active) parameters by choosing only top-$k$ experts for the inference of each token, thereby enhancing inference speed~\cite{sanseviero2023moe}. However, the static parameters, particularly those required for constructing the MoE architecture, still demand considerable memory and storage for deployment. For example, loading the Mixtral 8x7B model in \texttt{bf16} format requires at least two A100-80G GPUs. Notably, in this MoE model, the eight experts constitute around 96\% (45B out of 47B) of the total number of parameters. 

On the other hand, not all experts are equal in the MoE model. Recent studies, such as~\cite{chi2022on}, have demonstrated this discrepancy in expert training outcomes. The differing levels of training among each expert highlight the importance and practicality of identifying and pruning less significant experts, thereby improving the deployment efficiency of MoE models.

Unlike existing post-training weight pruning schemes for LLMs, which primarily target unstructured sparsity and N:M semi-structured sparsity~\cite{sun2023simple,frantar2023sparsegpt}, our approach focuses on expert-level sparsity for model sparsification. The aforementioned fine-grained weight pruning techniques are effective in reducing the total number of parameters. However, they face challenges in plug-and-play deployment due to the necessity for specific hardware designs (e.g., FPGA)~\cite{zhou2021learning}, which demands an extensive co-design of hardware and systems.

In this paper, we systematically explore \textit{expert-level} sparsity in MoE LLMs and, for the first time to our best knowledge, introduce hardware-friendly post-training methods for either permanently removing unimportant experts (expert pruning) or dynamically skipping experts during inference (dynamic expert skipping). Our proposed method significantly reduces memory usage for deploying MoE LLMs and enhances their inference speed.

\begin{figure*}[t!]
    \centering
    \includegraphics[width=\linewidth]{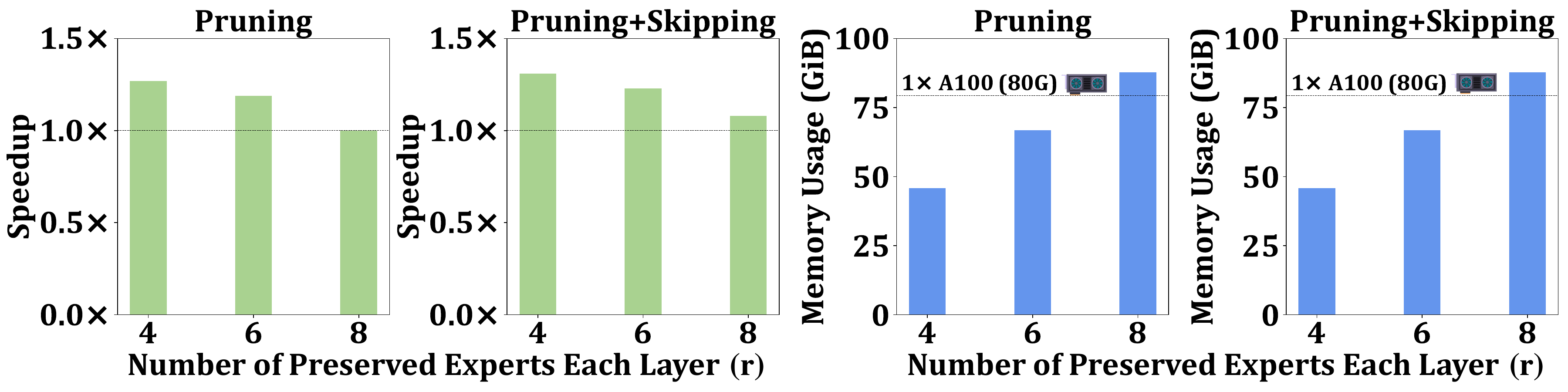}
    \vspace{-1em}
    \caption{Memory usage reduction (\texttt{bf16}) and inference speedup illustration of our proposed post-training expert pruning and dynamic (expert) skipping methods on the Mixtral 8x7B~\cite{jiang2024mixtral} model. 
    Our method greatly reduces memory consumption and enhances inference speed.}
    \label{fig:memory_speed}
    \vspace{-0.5em}
\end{figure*}

Initially, we investigate how to prune less important experts while maintaining satisfactory performance, utilizing an efficient post-training approach. We aim to minimize the token reconstruction loss in a layer-by-layer manner. Given the limited number of experts in a single MoE layer of the LLM, we meticulously enumerate and choose combinations of experts that yield the lowest token reconstruction loss, subsequently, concatenating them to obtain the final pruned MoE model. This strategy significantly lowers the memory demands for deploying MoE LLMs. We examine expert-level pruning for both task-agnostic and task-specific (first in literature) models, tailoring our strategies to optimize performance across a wide range of applications.

Building on this foundation, we further dive into strategies for accelerating the inference speed of MoE LLMs without compromising their robustness. Specifically, based on the model's fixed expert count, we introduce an online method for dynamically skipping certain experts. This approach, which is complementary to our expert pruning strategy, allows for on-the-fly adjustment of the number of active experts during inference, thus enhancing the inference speed. By integrating the dynamic (expert) skipping approach with expert pruning, we achieve a more streamlined and efficient deployment for MoE LLMs.

Experiments on Mixtral 8x7B (Instruct) models~\cite{jiang2024mixtral} demonstrate that our methods significantly reduce memory usage and increase inference speed. Take the example of post-training pruning two experts. As shown in Fig.~\ref{fig:memory_speed}, we halve the number of GPUs needed, allowing deployment on a single 80G GPU and achieving a 1.2$\times$ inference speedup. The pruning also results in mild performance loss, specifically, around 2.9 points for task-agnostic and 6.2 points (reducible to 1.6 with task-specific fine-tuning) for task-specific models. Further combination of dynamic skipping with expert pruning can lead to the same inference speedup with dropping 4 experts while achieving much higher model performances. To the best of our knowledge, this study is the first to discuss expert-level sparsity and propose efficient schemes for expert pruning and skipping for MoE LLMs.

\section{Related Works}\label{related_works}

\subsection{Mixture-of-Experts Models}
First introduced in~\cite{jacobs1991adaptive}, a Mixture-of-Experts (MoE) model contains multiple separate networks, and each network processes a subset of the entire dataset. This separation can be viewed as a modular transformation of a multi-layer network. MoE structure is used for designing Recurrent Neural Networks (RNNs) in~\cite{shazeer2017outrageously} and further extended to encoder-decoder Transformer-based models~\cite{lepikhin2020gshard}. With the recent development of decoder-only GPT family of models~\cite{brown2020language,touvron2023llama1,touvron2023llama2}, MoE models based on this structure gain popularity~\cite{jiang2024mixtral}. In this paper, we focus on post-training expert pruning/skipping methodologies for MoE LLMs.

\subsection{Expert Pruning for MoE Models}
Expert pruning within MoE models has garnered attention in the realm of Natural Language Processing, particularly in machine translation tasks. In these contexts, the translation of specific languages often renders the expertise of other language specialists superfluous. The most activated experts are reserved in~\cite{kim2021scalable} to prune a machine translation MoE model, and~\cite{koishekenov2022memory} proposes expert pruning metrics based on gate statistics collected during decoding. Although these methods actively deal with expert pruning for MoE models, they are still limited to the machine translation domain with linguistic models. Researchers in~\cite{chen2022task} provide a dropping-while-training method that progressively drops the non-professional experts for target downstream tasks, and experiments are carried out on Switch Transformers models~\cite{fedus2022switch}. However, in the LLM era, it is usually difficult to afford such a training paradigm.

\subsection{Post-training Pruning for LLMs}
Post-training pruning~\cite{kwon2022fast,hubara2021accelerated} has become a popular topic for neural network sparsification in recent years. Given a trained model, post-training pruning aims at achieving the optimal model sparsification outcome by utilizing model parameters together with some calibration data. Recent works extend pruning methods to LLMs~\cite{sun2023simple,frantar2023sparsegpt}. However, these pruning methods primarily focus on sparsifying the weight matrices of linear layers in the LLMs and require dedicated hardware. To the best of our knowledge, efficient post-training expert pruning methods have not been discussed for decoder-only LLMs with MoE structures.

\section{Method}\label{method}
To enhance the deployment efficiency of MoE LLMs, we concentrate on expert-level model sparsity and innovatively propose post-training techniques designed to reduce memory usage and increase inference speed. In this section, we offer a comprehensive explanation of our proposed methods for expert pruning (Sec.~\ref{sec:method}) and dynamic (expert) skipping (Sec.~\ref{sec:online_method}), considering both memory consumption and inference speed. 

\begin{figure}[t!]
    \centering
    \includegraphics[width=1.01\linewidth]{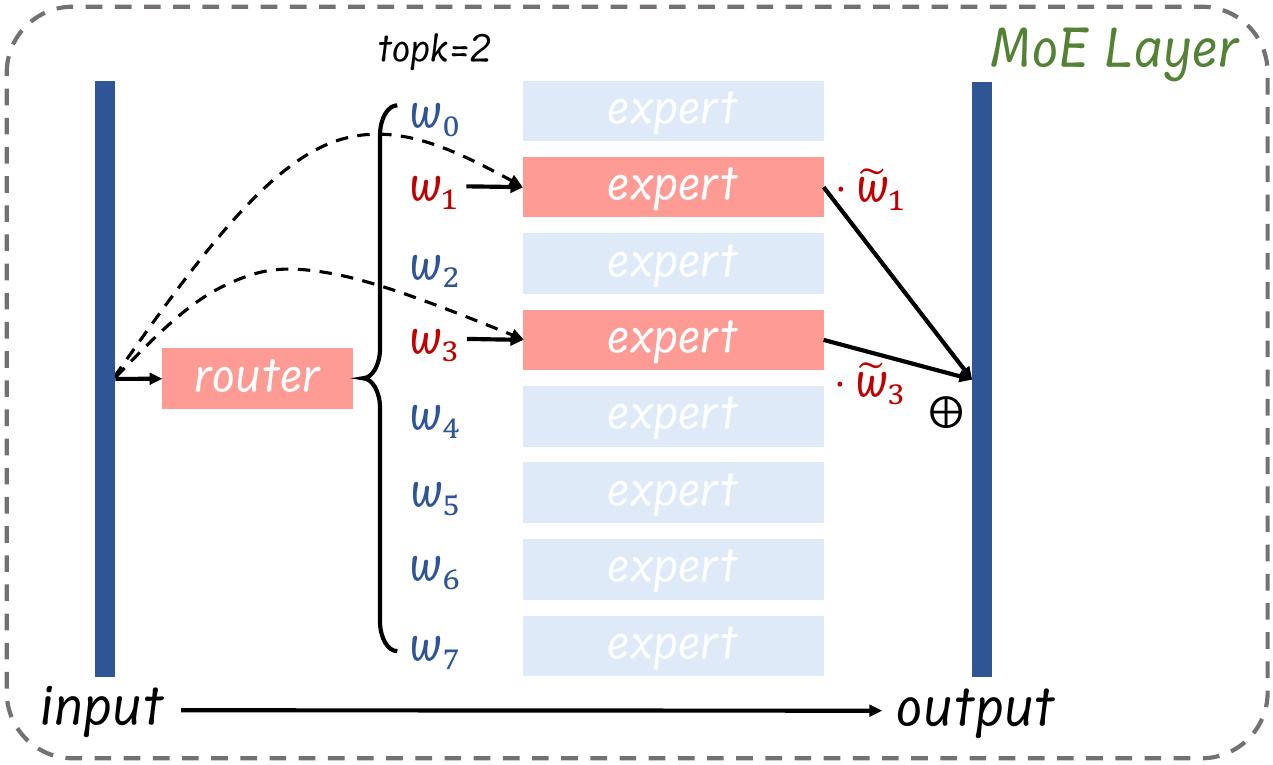}
    \caption{Illustration of the MoE layer in the Mixtral 8x7B model for per-token inference. The output of the layer is the weighted sum of the outputs from selected experts over input token $\boldsymbol{x}$. $\widetilde{w}_i$ denotes the normalized routing weight of each selected expert.}
    \label{fig:per_token_moe}
    \vspace{-1em}
\end{figure}

\subsection{Preliminary}\label{sec:preliminary}

In the decoder-only sparse MoE Transformer models, as discussed in~\cite{gale2023megablocks}, the Feed-Forward Network (FFN) sub-layers of the traditional dense model are replaced with MoE layers, each containing $n$ experts. Specifically, within the MoE layer of the Mixtral 8x7B model, featuring 8 experts as detailed in~\cite{jiang2024mixtral}, each token $\boldsymbol{x}$ in the input sequence is routed to the top-2 experts based on the routing weights $\boldsymbol{w}$.

\begin{figure*}[t]
    \centering
    \includegraphics[width=\textwidth]{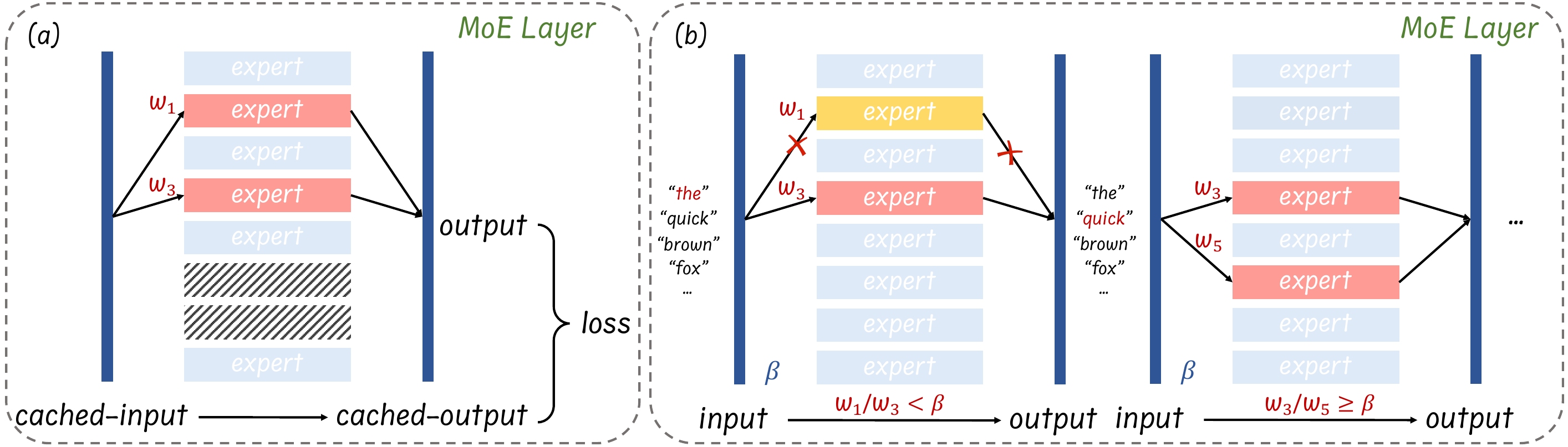}
    \vspace{-1em}
    \caption{Framework of our proposed expert pruning and dynamic skipping methods. (a) The expert pruning method evaluates the contributions of experts via a small calibration dataset and then permanently discards those with low contributions (e.g., experts with a slashed background). (b) The dynamic skipping method discard no experts instead dynamically decides whether to skip certain experts (e.g., experts with a yellow background) during inference.}
    \label{fig:drop_moe}
\end{figure*}
\begin{figure*}[t]
    \centering
    \includegraphics[width=\textwidth]{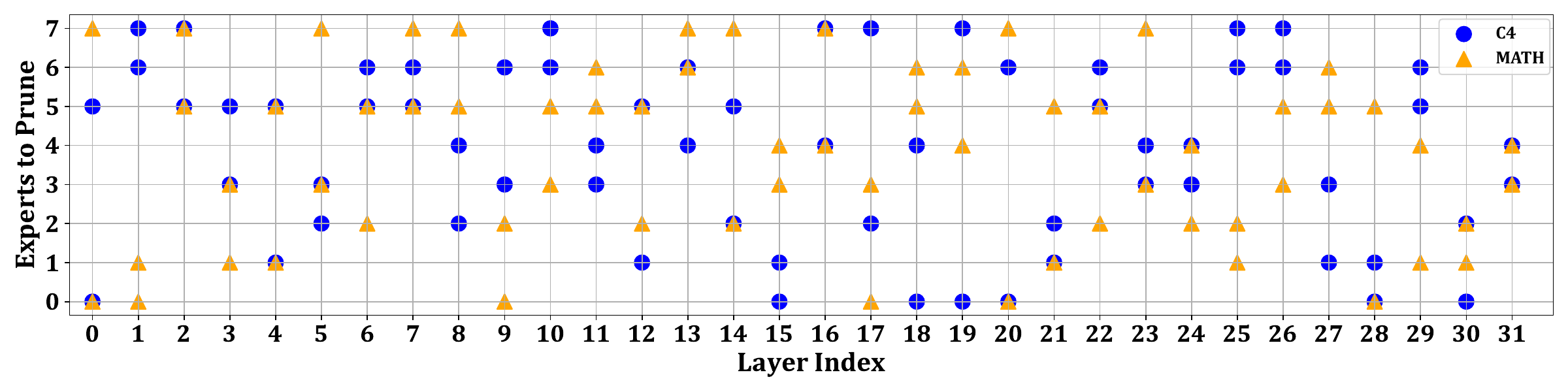}
    \vspace{-0.5em}
    \caption{Expert selection comparison between C4 and MATH dataset with $r=6$ for Mixtral 8x7B model. Significant divergence is observed in the selection of experts across these two datasets, and identical expert combinations are observed in only four specific layers (i.e., layer 2, layer 4, layer16, and layer 31).}
    \label{fig:c4_math_expert}
    \vspace{-0.5em}
\end{figure*}

The inference process for each input token $\boldsymbol{x}$ within the MoE layer of the Mixtral 8x7B decoder layer is depicted in Fig.~\ref{fig:per_token_moe}. Initially, the router computes routing logits $\boldsymbol{l}=\{{l}_0,\dots, {l}_{n-1}\}$ and routing weights $\boldsymbol{w} = \text{Softmax}(\boldsymbol{l})$ for the experts. Then the top-$k$ experts, where $k=2$ for the Mixtral 8x7B model, are selected based on their routing weights to process the token. Each of these $k$ selected experts, applying a SwiGLU transformation $\mathcal{E}_{i}(\cdot)$ ($i\in\{0, 1, \dots, n-1\}$), contributes to the final output. This output is a weighted sum of the individual expert outputs, with weights $\widetilde{{w}}_i$ being the normalized values of the corresponding routing weights for the selected experts. The normalized weight for expert $e_j$ ($j\in\{0, 1, \dots, k-1\}$)\footnote{$e_j$ is the index of the $j$-th largest element of $\boldsymbol w$, i.e. the index of the $j$-th selected expert} is calculated as follows:
\begin{align}
\widetilde{{w}}_{e_j}=\frac{{w}_{e_j}}{\sum_{m=0}^{k-1}{w}_{e_m}},
\end{align}
yielding the MoE layer's output for the token $\boldsymbol{x}$ as:
\begin{align}
\boldsymbol{z}=\sum_{j=0}^{k-1}\widetilde{{w}}_{e_j}\cdot \mathcal{E}_{e_j}(\boldsymbol{x}).
\end{align}
Aside from these specified mechanisms, the remaining aspects of the MoE network mirror those of a standard decoder-only Transformer model. 

\subsection{Post-training Expert Pruning}\label{sec:method}

As demonstrated in the preliminary subsection, the parameters of experts occupy the major proportion of the whole MoE LLM model. However, for a single token, only a small subset of these experts are activated, leading to considerable inefficiencies in parameter utilization. Existing post-training weight pruning methods for LLMs (e.g., Wanda~\cite{sun2023simple}, SparseGPT~\cite{frantar2023sparsegpt}), while effective in reducing model parameters, do not support efficient deployment of MoE LLMs without specialized hardware implementations~\cite{mishra2021accelerating}. Therefore, we introduce a heuristic search method to prune the number of experts in a post-training manner.

\textbf{Task-agnostic Expert Pruning for General Tasks.} Different from existing pruning schemes leveraging unstructured or semi-structured weight sparsity~\cite{sun2023simple,frantar2023sparsegpt} for LLMs, our proposed post-training expert pruning method aims at reducing the parameter numbers of MoE LLMs by permanently discarding less important experts, thereby improve the inference efficiency. It is a post-training pruning method and does not require any parameter update.
Fig.~\ref{fig:drop_moe} (a) illustrates our proposed post-training expert pruning method. We conduct expert pruning in a layer-wise manner. Specifically, the pruning method contains two steps as follows. 

Firstly, we set up a small calibration dataset, then perform inference on the original MoE model with all experts over the dataset, and cache the input-output token pairs of each MoE layer. We use samples from the pre-training dataset C4~\cite{2019t5} as calibration data, since pre-training datasets are often more comprehensive and not dominated by knowledge specific to any particular domain~\cite{sun2023simple}. 

Secondly, after caching input-output pairs for each MoE layer, we enumerate expert combinations based on the preset parameter $r$, denoting the number of preserved experts. Assume the function of the MoE layer at layer $l$ is $\mathcal{F}(\cdot)$, with the cached input represented by $\boldsymbol{x}$. During each enumeration, we maintain $r$ experts and eliminate the remaining experts along with their associated routing weights. Subsequently, we employ the pruned MoE layer, denoted as $\mathcal{F}'(\cdot, \mathbf{C})$, to recalculate the corresponding output, where $\mathbf{C}$ represents a subset containing $r$ experts selected from the original $n$ experts. Inspired by channel pruning in CNNs~\cite{he2017channel}, the Frobenius norm of the difference between cached output $\mathcal{F}(\boldsymbol{x})$ and 
the output of pruned layer $\mathcal{F}'(\boldsymbol{x},\mathbf{C})$ is used to quantify the discrepancy between the model before and after expert pruning, and we denote it as \textit{reconstruction loss}. The expert subset corresponding to the minimum reconstruction loss is retained. The $n-r$ experts left are considered to contribute least to the original MoE model and are thus discarded. The expert pruning process in layer $l$ can be formulated as:
\vspace{-0.4em}
\begin{align}
    \min_\mathbf{C}&\|\mathcal{F}'(\boldsymbol{x},\mathbf{C})-\mathcal{F}(\boldsymbol{x})\|_F\\
    \text{s.t. }\mathbf{C}&\subseteq\{\text{expert}_0,\dots, \text{expert}_{n-1}\},|\mathbf{C}|=r.\nonumber
\end{align}
We heuristically search for expert subset $\mathbf{C}$ with the lowest reconstruction loss in each layer separately and obtain an MoE model with $r$ experts by the concatenation of each pruned layer. After removing the insignificant experts, the pruned model can be easily loaded using existing packages (e.g., Huggingface Transformers~\cite{wolf2020transformers}) with just a change of the model configuration. Especially, with 2 experts pruned, the deployment budget is reduced to a single 80G GPU for loading the Mixtral 8x7B (Instruct) model with \texttt{bf16} data type.

\textbf{Task-specific Expert Pruning for Domain-specific Tasks.} Previous research on post-training pruning for LLMs~\cite{sun2023simple,frantar2023sparsegpt} typically considers the performance over general tasks. For the first time, in our work, we investigate the task-specific post-training pruning for MoE LLMs. Our above-proposed expert pruning strategy is adept at conserving the knowledge encapsulated by the experts of an MoE model for general tasks. However, as a pre-training dataset, C4 spans a wide array of domains, posing challenges when pruning experts for domain-specific tasks (e.g., models tailored for mathematics~\cite{yu2023metamath,wang2024mathcoder}). We evaluate the 5-shot performance of the C4 pruned MoE LLM model with $r=6$ on math tasks (GSM8K)~\cite{cobbe2021training}, resulting in a performance degradation plummeting from 58.61 to 41.02 for the Mixtral 8x7B model. To address this challenge, we propose to shift the calibration dataset \textit{from C4 to the training set of MATH}~\cite{hendrycksmath2021}, to concentrate the pruning process on the mathematics domain. 

\textbf{Remark.} For a better comparison between general tasks and domain-specific tasks, we visualize the distribution of pruned experts selected by C4 and MATH, as shown in Fig.~\ref{fig:c4_math_expert}. For both the Mixtral 8x7B and Mixtral 8x7B Instruct model, identical expert combinations are observed in only four specific layers. This suggests that there are distinct differences in the distributions of pre-training datasets and domain-specific datasets. Further details and discussions are provided in Sec.~\ref{sec:specific}.

\begin{table*}[t]\huge
\vspace{-1em}
\renewcommand\arraystretch{1.1}
\resizebox{\textwidth}{!}{%
\begin{tabular}{clcccccccccccc}
\midrule
\textbf{Model} & \textbf{Method} & \textbf{Sparsity}& \textbf{ARC-c} & \textbf{ARC-e} & \textbf{BoolQ} & \textbf{HellaSwag} & \textbf{MMLU} & \textbf{OBQA} & \textbf{RTE} & \textbf{WinoGrande} & \textbf{Average} & \textbf{Mem (MB)} &\textbf{Speedup} \\ \midrule
\multirow{2}{*}{\textbf{Mixtral 8x7B}}          & \textbf{Wanda}  & 2:4      & 42.06          & 74.16     & 76.64 & 53.16     & 52.21 & 27.00      & 63.90 & 70.96      & 57.51   & 51,214      & 0.91$\times$    \\\cmidrule{2-14}
& \textbf{Ours}   & $r=4$   & 48.89          & 78.16     & 81.35 & 57.66     & 47.30 & 29.00      & 61.37 & 72.85      & \textbf{59.57}   & \textbf{46,879}      & \textbf{1.27}$\times$    \\ \midrule
\multirow{2}{*}{\textbf{\makecell{Mixtral 8x7B\\ Instruct}}} & \textbf{Wanda}  & 2:4      & 48.89          & 78.70     & 86.27 & 56.24     & 57.84 & 30.40      & 72.20 & 71.82      & 62.80   & 51,210      & 0.92$\times$    \\\cmidrule{2-14}
& \textbf{Ours}   & $r=4$   & 53.92          & 79.88     & 84.77 & 60.05     & 52.75 & 30.40      & 75.45 & 73.80      & \textbf{63.88}   & \textbf{46,879}      & \textbf{1.27}$\times$   \\ \midrule
\end{tabular}
}
\vspace{-0.2em}
\caption{Comparison with Wanda~\cite{sun2023simple} at the structured 2:4 sparsity pattern. Our proposed expert pruning method ($r=4$) outperforms Wanda in all aspects. \textbf{Mem} stands for memory usage. $^\dagger$The original average performance of Mixtral 8x7B and Mixtral 8x7B Instruct model is 67.58 and 69.98, respectively.}
\vspace{-0.2em}
\label{tab:cmp_wanda}
\end{table*}

\begin{table*}[t]\small
\resizebox{\textwidth}{!}{%
\begin{tabular}{clcccccccccc}
\midrule
\textbf{Model} &
  \textbf{Method} &
  \textbf{$r$} &
  \textbf{ARC-c} &
  \textbf{ARC-e} &
  \textbf{BoolQ} &
  \textbf{HellaSwag} &
  \textbf{MMLU} &
  \textbf{OBQA} &
  \textbf{RTE} &
  \textbf{WinoGrande} &
  \textbf{Average} \\ \midrule
\multirow{7}{*}{\textbf{Mixtral 8x7B}} &
  \textbf{None} &
  8 &
  57.17 &
  84.01 &
  85.35 &
  64.88 &
  67.88 &
  35.00 &
  70.40 &
  75.93 &
  67.58 \\ \cmidrule{2-12} 
 & \multirow{2}{*}{\textbf{Random}}   & 6 & 48.04 & 78.49 & 81.99 & 59.02 & 60.77 & 33.40 & 66.79 & 75.85 & 63.04 \\ 
 & & 4 & 39.85 & 68.35 & 78.59 & 53.32 & 49.23 & 29.20 & 62.82 & 69.93 & 56.41 \\ \cmidrule{2-12} 
 & \multirow{2}{*}{\textbf{Frequency}} & 6 & 49.06 & 78.83 & 77.03 & 59.38 & 55.18 & 33.60  & 57.40  & 75.69 & 60.77 \\ 
 &  & 4 & 43.86 & 73.61 & 76.97 & 54.01 & 41.48 & 26.20  & 57.04 & 73.48 & 55.83 \\ \cmidrule{2-12} 
 & \multirow{2}{*}{\textbf{Ours}}     & 6 & 51.62 & 81.94 & 83.64 & 61.60 & 58.72 & 33.00 & 67.87 & 75.37 & \textbf{64.22} \\ 
 & & 4 & 48.89 & 78.16 & 81.35 & 57.66 & 47.30 & 29.00 & 61.37 & 72.85 & \textbf{59.57} \\ \midrule
\multirow{7}{*}{\textbf{\makecell{Mixtral 8x7B\\Instruct}}} &
  \textbf{None} &
  8 &
  62.20 &
  87.04 &
  88.50 &
  67.59 &
  68.87 &
  36.60 &
  72.20 &
  76.87 &
  69.98 \\ \cmidrule{2-12} 
 & \multirow{2}{*}{\textbf{Random}}   & 6 & 54.52 & 83.04 & 87.25 & 63.21 & 62.70 & 35.40 & 72.92 & 77.19 & 67.03 \\ 
 & & 4 & 48.81 & 75.46 & 78.47 & 57.48 & 53.68 & 31.80 & 72.56 & 70.96 & 61.15 \\ \cmidrule{2-12} 
 & \multirow{2}{*}{\textbf{Frequency}} & 6 & 55.89 & 82.83 & 86.33 & 63.69 & 58.89 & 37.00 & 63.18 & 76.01 & 65.48 \\ 
 & & 4 & 49.40 & 77.27 & 82.97 & 57.66 & 47.03 & 32.20 & 66.79 & 74.03 & 60.92 \\ \cmidrule{2-12} 
 & \multirow{2}{*}{\textbf{Ours}}     & 6 & 58.19 & 84.89 & 87.34 & 65.24 & 62.47 & 35.60 & 70.04 & 75.85 & \textbf{67.45} \\ 
 & & 4 & 53.92 & 79.88 & 84.77 & 60.05 & 52.75 & 30.40 & 75.45 & 73.80 & \textbf{63.88} \\ \midrule
\end{tabular}%
}
\vspace{-0.5em}
\caption{Zero-shot performance evaluation of different expert pruning methods with $r$ set to 6 and 4. \textbf{Random} stands for randomly choosing experts to discard in each MoE layer. \textbf{Frequency} stands for dropping experts based on their activation frequency during the inference over calibration data. Our proposed expert pruning method leads to the least performance drop, with around 2.9 points for dropping 2 experts and 7.1 points for dropping 4 experts.}
\vspace{-0.5em}
\label{tab:offline_general}
\end{table*}

\subsection{Dynamic Skipping During Inference}\label{sec:online_method}
Our expert pruning strategy effectively reduces memory consumption during model deployment. However, each token is still processed by $k$ selected experts, without a reduction in runtime FLOPs. Intuitively, not all tokens require the selection of all top-$k$ experts during the token generation process. Consequently, we introduce a scheme that dynamically skips certain experts for individual token inference, to further enhance inference efficiency.

As shown in Fig.~\ref{fig:per_token_moe}, during the inference process, top-$k$ experts are chosen with routing weights ${\boldsymbol{w}}=\{w_{e_0}, w_{e_1},\dots,w_{e_{k-1}}\}$ for each token $\boldsymbol{x}$ in an MoE layer. For simplicity, we assume $k=2$ (as in Mixtral 8x7B)\footnote{Deeper theoretical insight and a broader application to the top-$k$ scenario of dynamic skipping can be found in Sec.~\ref{sec:theory_dynamic_skipping}.}. Our proposed dynamic expert skipping method is illustrated in Fig.~\ref{fig:drop_moe} (b). Without loss of generality, assume experts with indices $e_0$ and $e_1$ are chosen, and $w_{e_1}<w_{e_0}$. To accelerate inference speed, if $w_{e_1}<\beta w_{e_0}$, we do not assign $\boldsymbol{x}$ to expert $e_1$, where $\beta$ is a hyper-parameter separately set for each MoE layer. In our implementation, we forward the model over the sampled calibration data and set $\beta$ as the median value of $\frac{w_{e_1}}{w_{e_0}}$ for each MoE layer. The dynamic skipping scheme can be carried out on the fly to speed up inference, and can be used simultaneously with expert pruning. In our experiments, we observe a 1.2$\times$ to 1.3$\times$ inference speedup with $r=6$.

\section{Experiment}\label{experiment}

In this section, a series of experiments are carried out to evaluate our proposed methods. We introduce experiments of expert pruning for general tasks in Sec.~\ref{sec:general}, domain-specific tasks in Sec.~\ref{sec:specific}, and dynamic expert skipping results in Sec.~\ref{sec:online}.

\subsection{Expert Pruning for General Tasks}\label{sec:general}

In this subsection, we evaluate the proposed expert pruning method on some general tasks, which can comprehensively reflect the knowledge retention of the model after expert pruning.

\textbf{Experiment Setup. }Similar to Wanda~\cite{sun2023simple}, we choose calibration data from the C4~\cite{2019t5} dataset and combine them into 128 token sequences, each with a length of 2048\footnote{For more experiments about the influence of the calibration dataset size, please refer to Sec.~\ref{sec:calib_size}.}. We perform expert pruning on both Mixtral 8x7B and Mixtral 8x7B Instruct models, resulting in MoE models with two experts discarded ($r=6$) and four experts discarded ($r=4$) in each layer. Pruning a Mixtral 8x7B model takes about 30 minutes for $r=6$ and 90 minutes for $r=4$.

After expert pruning, we evaluate the performance of the pruned MoE models following Wanda~\cite{sun2023simple}. Specifically, we report zero-shot accuracies of 8 tasks from EleutherAI LM Harness~\cite{eval-harness}. We also test the token generation speed\footnote{We revise the script provided in \url{https://github.com/AutoGPTQ/AutoGPTQ/} to test token generating speed. 
}, together with the peak GPU memory usage during model inference.

\textbf{Comparison with Weight Pruning Methods.} We compare our proposed expert pruning method with the representative weight pruning algorithm Wanda. For a fair comparison, we set $r=4$ in our method and test Wanda with the commonly used structured 2:4 sparsity pattern. This will lead to around 50\% parameter reduction for both methods. The results are shown in Tab.~\ref{tab:cmp_wanda}. The inference speedup of 2:4 structured model relies on specially designed hardware~\cite{mishra2021accelerating} and scripts. In our experiments, we even observe a lower inference speed compared with the dense weight model\footnote{Our implementation is based on \url{https://pytorch.org/tutorials/prototype/semi_structured_sparse.html}.}. Besides, our expert pruning method excels Wanda with the 2:4 sparsity pattern in both memory usage and benchmark performance. 

\textbf{Comparison with Other Expert Pruning Baselines.} We also set up two baseline methods. One baseline is randomly dropping experts in each layer. The other method calculates the activation frequency of each expert during forward passes on the calibration data and discards those with the lowest activation frequencies in each layer. Comparison results are listed in Tab.~\ref{tab:offline_general}. The method based on activation frequency gets the worst performance. This phenomenon implies that although the MoE model might show a tendency for expert selection during the inference process, simply carrying out expert pruning based on activation frequency might not always lead to desirable results. In contrast, our proposed method achieves better results. Compared to the origin model with 8 experts, our model achieves a 2.9-point performance drop with 24\% parameter reduction and a 7.1-point performance drop with 48\% parameter reduction on average without any extra training. 

\textbf{Memory Usage and Generation Speed.} The memory usage statistics are shown in Fig.~\ref{fig:memory_speed}. It takes 2 A100-80G GPUs to load and forward the original 8-expert model with \texttt{bf16} data type. After pruning 2 and 4 experts, only one 80G GPU is needed for the inference process. For token generation speed analysis, during model inference, we still need to route each token to two experts. However, reducing the number of GPUs required to load the model can decrease the time consumed by GPU intercommunication, resulting in a much higher token generation speed. We observe a 1.20$\times$ token generation speedup for the model with 2 experts pruned and a 1.27$\times$ speedup with 4 experts pruned.

\subsection{Expert Pruning for Domain-Specific Tasks}\label{sec:specific}

\begin{table}[t]
\centering
\resizebox{0.98\columnwidth}{!}{%
\begin{tabular}{clcc}
\midrule
\textbf{Model} & \textbf{Method} & \textbf{Sparsity} & \textbf{GSM8K (5-shot)} \\ \midrule
\multirow{7}{*}{\textbf{Mixtral 8x7B}} & \text{None} & {None} & 58.61 \\\cmidrule{2-4}
& Wanda \textbf{(C4)} & 2:4 & 14.10 \\
& Wanda \textbf{(MATH)} & 2:4 & \textbf{20.39} \\\cmidrule{2-4}
& Random & $r=4$ & 0.68 \\
& Ours \textbf{(C4)} & $r=4$ & 24.87 \\
& Ours \textbf{(MATH)} & $r=4$ & \textbf{37.07} \\\cmidrule{2-4}
& Random & $r=6$ & 36.39 \\
& Ours \textbf{(C4)} & $r=6$ & 41.02 \\
& Ours \textbf{(MATH)} & $r=6$ & \textbf{51.25} \\\midrule
\multirow{7}{*}{\textbf{\makecell{Mixtral 8x7B\\Instruct}}} & \text{None} & {None} & 63.46 \\\cmidrule{2-4}
& Wanda \textbf{(C4)} & 2:4 & 26.69 \\
& Wanda \textbf{(MATH)} & 2:4 & \textbf{31.31} \\\cmidrule{2-4}
& Random & $r=4$ & 0.76 \\
& Ours \textbf{(C4)} & $r=4$ & 30.40 \\
& Ours \textbf{(MATH)} & $r=4$ & \textbf{47.01} \\\cmidrule{2-4}
& Random & $r=6$ & 39.80 \\
& Ours \textbf{(C4)} & $r=6$ & 48.52 \\
& Ours \textbf{(MATH)} & $r=6$ & \textbf{58.38} \\\midrule
\end{tabular}
}
\vspace{-5pt}
\caption{5-shot GSM8K~\cite{cobbe2021training} accuracy comparison for sampling calibration data from different datasets. Random pruning will lead to bad performance in this case. Compared with pre-training datasets, sampling from domain-specific datasets will significantly improve the performance on corresponding tasks after expert pruning. Our expert pruning strategy also outperforms Wanda with 2:4 structured sparsity.}
\vspace{-1.4em}
\label{tab:gsm8k_few}
\end{table}

\textbf{Experiment Setup.} We investigate on mathematical reasoning tasks. We randomly sample sentences from the train set of MATH~\cite{hendrycksmath2021} and combine them into 128 token sequences, each with a length of 2048. We carry out expert pruning with $r=6$ and $r=4$, then test 5-shot GSM8K~\cite{cobbe2021training} results.

\textbf{Baselines to Compare.} We compare the 5-shot GSM8K result with randomly pruned models, models pruned using samples from C4 as calibration data, the 2:4 structured model obtained by Wanda, as well as the original MoE model with 8 experts. Vanilla Wanda uses C4 for data calibration. To test the influence of the calibration dataset on model performance, we also leverage the MATH dataset for Wanda pruning.

\textbf{Evaluation Results.} Tab.~\ref{tab:gsm8k_few} illustrates the 5-shot evaluation results on the GSM8K dataset. The performance witnesses a significant drop after random pruning, or pruning with calibration data obtained from C4. However, this degradation dramatically reduces after leveraging the MATH dataset for calibration data construction. A similar phenomenon is also observed with Wanda. This indicates that when facing domain-specific tasks, using datasets corresponding to these specific tasks can yield better expert pruning results than using pre-training datasets. It also implies that our proposed method for changing the calibration dataset for domain-specific tasks can also be applied to other pruning algorithms. Besides, our expert pruning strategy ($r=4$) significantly outperforms Wanda with 2:4 structured sparsity.

\begin{table}[t]\small
\centering
\resizebox{0.98\columnwidth}{!}{%
\begin{tabular}{clccc}
\midrule
\textbf{Model} & \textbf{Method}& \textbf{$r$} & \textbf{GSM8K} & \textbf{MATH} \\ \midrule
\textbf{MetaMath 70B} & \text{None}& N/A & 82.30 & 26.60 \\ \midrule
\multirow{4}{*}{\textbf{Mixtral 8x7B}} & \text{None}  & 8& 81.35 & 34.86\\ \cmidrule{2-5} 
& Ours \textbf{(C4)}& 6& 79.53 & 32.48\\ 
& Ours \textbf{(MATH)} & 6& 79.53 & 33.58\\ 
& Ours \textbf{(MATH)} & 7& \textbf{81.20} & \textbf{34.40}\\ \midrule
\multirow{4}{*}{\textbf{\makecell{Mixtral 8x7B\\Instruct}}} & \text{None}  & 8& 81.43 & 35.46\\ \cmidrule{2-5} 
& Ours \textbf{(C4)}& 6& 79.83 & 32.70\\ 
& Ours \textbf{(MATH)} & 6& 80.06 & 34.10\\ 
& Ours \textbf{(MATH)} & 7& \textbf{81.50} & \textbf{34.86}\\ \midrule
\end{tabular}%
}
\vspace{-5pt}
\caption{Zero-shot evaluation results of GSM8K~\cite{cobbe2021training} and MATH~\cite{hendrycksmath2021} after training the MoE models on MetaMathQA~\cite{yu2023metamath} with different expert numbers and different calibration datasets for expert pruning. Using domain-specific data can result in better performance on corresponding downstream tasks. Model fine-tuning can greatly reduce the performance gaps between the pruned models and the original model.}
\vspace{-1.4em}
\label{tab:gsm8k_math}
\end{table}

\textbf{More Discussion. }Using samples from the MATH dataset can greatly improve the domain-specific performance on mathematics tasks. However, as our method follows a post-training manner without any training, the expert pruning scheme still leads to a significant performance drop. To reduce the performance gap between pruned models and original models, we fully fine-tune the MoE models with different expert numbers on the MetaMathQA~\cite{yu2023metamath} dataset and compare their performances. We fine-tune and compare models pruned by C4 ($r=6$), MATH ($r=6, r=7$), and the original 8-expert model. The zero-shot GSM8K@1 and MATH@1 accuracies after fine-tuning different MoE models are shown in Tab.~\ref{tab:gsm8k_math}. As can be seen, the fine-tuning process significantly reduces the performance drop incurred by expert pruning and leads to comparable results with tuning full-expert models. Specifically, for the Mixtral 8x7B Instruct model, the accuracy of the pruned 7-expert model on the GSM8K test set exceeds that of the 8-expert model. This suggests that for certain practical downstream tasks, a large number of experts might not be a necessity for achieving good performance. Also, the pruned models using samples from MATH as the calibration dataset outperform those using C4 after tuning, further highlighting the effectiveness of adopting domain-specific calibration datasets for task-specific models.

\begin{table}[t]\small
\centering
\resizebox{0.98\columnwidth}{!}{%

\begin{tabular}{cccccc}
\midrule
\textbf{Model} & $r$ & \textbf{Pruning} & \textbf{Skipping} & \textbf{LM-eval} & \textbf{Speedup} \\ \midrule
\multirow{6}{*}{\textbf{Mixtral 8x7B}}  & 8 & & & 67.58 & 1.00$\times$ \\\cmidrule{2-6}
 & 8 & &\checkmark & \textbf{66.37} & \textbf{1.08$\times$} \\
 & 6 &\checkmark & & 64.22 & 1.19$\times$ \\
 & 6 &\checkmark &\checkmark & \textbf{62.91} & \textbf{1.23$\times$} \\
 & 4 &\checkmark & & 59.57 & 1.27$\times$ \\
 & 4 &\checkmark &\checkmark & \textbf{57.91} & \textbf{1.31$\times$} \\
 \midrule
\multirow{6}{*}{\textbf{\makecell{Mixtral 8x7B\\ Instruct}}} & 8 & & & 69.98 & 1.00$\times$ \\\cmidrule{2-6}
 & 8 & &\checkmark & \textbf{69.03} & \textbf{1.08$\times$} \\
 & 6 &\checkmark & & 67.45 & 1.20$\times$ \\
 & 6 &\checkmark &\checkmark & \textbf{66.04} & \textbf{1.27$\times$} \\
 & 4 &\checkmark & & 63.88 & 1.27$\times$ \\
 & 4 &\checkmark &\checkmark & \textbf{62.33} & \textbf{1.33$\times$} \\
 \midrule
\end{tabular}
}
\vspace{-5pt}
\caption{Evaluation results of combining expert pruning with dynamic skipping. We carry out expert pruning using calibration data sampled from C4, then infer the pruned models with dynamic expert skipping. We set $\beta$ as the median value of ${w}_{e_1}/{w}_{e_0}$ of the calibration set. The dynamic expert skipping method further enhances inference speed with a slight performance drop.}
\vspace{-1.5em}
\label{table:online_only}
\end{table}

\subsection{Dynamic Expert Skipping Results}\label{sec:online}
This subsection assesses the effectiveness of our proposed dynamic expert skipping approach. Additionally, we explore the combination of both methods to further improve inference efficiency.

\textbf{Experiment Setup.} We perform tests on task-agnostic models for better representativeness. The setup is similar to Sec.~\ref{sec:general}. We first prune the Mixtral 8x7B and Mixtral 8x7B Instruct model using calibration data sampled from the C4 dataset and get the pruned models with $r=6$ and $r=4$. During the testing of different benchmarks, we dynamically skip certain experts. For evaluation, we report the zero-shot accuracies of 8 tasks from EleutherAI LM Harness~\cite{eval-harness}. Our proposed dynamic expert skipping method does not influence the memory usage for model inference, so we just report the inference speed in this subsection.

\textbf{Baselines to Compare.} We suggest that our proposed dynamic expert skipping can be seamlessly integrated with the expert pruning approach. For setting up baselines, we evaluate zero-shot accuracies of the original 8-expert model without dynamic expert skipping, as well as the accuracies of models pruned to $r=6$ and $r=4$ without dynamic expert skipping. Subsequently, we incorporate dynamic expert skipping into the inference process of these models, evaluate accuracies over benchmarks, and measure token generation speedup.

\textbf{Evaluation Results.} The evaluation results are illustrated in Tab.~\ref{table:online_only}. We show the average accuracies of the 8 zero-shot tasks at the ``LM-eval'' column, together with the inference speedup ratio compared to the original 8-expert models. As can be seen, based on expert pruning, the dynamic expert skipping method can further enhance the inference speed with just negligible performance drops. We can achieve nearly 90\% performance of the Mixtral 8x7B Instruct model with half parameters and a 1.33$\times$ token generation speedup. Another notable observation is that the Mixtral 8x7B Instruct model using both expert pruning and dynamic skipping with $r=6$ achieves the same inference speedup as the model using only expert pruning with $r=4$ while getting a much higher accuracy over the LM-eval benchmark. This phenomenon also proves the efficiency of our dynamic expert skipping approach. For a more comprehensive evaluation, we perform dynamic skipping on task-specific models and observe similar results. Experiment details are shown in the Appendix (Sec.~\ref{sec:skip_domain}).

\begin{table}[t]\small
\centering
\resizebox{0.85\columnwidth}{!}{%
\renewcommand\arraystretch{0.65}
\begin{tabular}{clcc}
\midrule
\textbf{Model}& \textbf{Method} & \textbf{$r$} & \textbf{LM-eval} \\ \midrule
\multirow{4}{*}{\textbf{Mixtral 8x7B}} & Progressive & 6& \textbf{64.48} \\ 
 & Layer-wise (\textbf{Ours}) & 6 & 64.22\\ \cmidrule{2-4} 
 & Progressive & 4 & 57.53\\ 
 & Layer-wise (\textbf{Ours})& 4 & \textbf{59.57} \\ \midrule
\end{tabular}%
}
\vspace{-5pt}
\caption{Comparison between the layer-wise pruning manner and the progressive pruning manner. The progressive scheme will lead to more performance degradation with a high expert pruning rate.}
\label{tab:ablation}
\vspace{-1.5em}
\end{table}

\subsection{More Analysis}

\textbf{Discussion about Inference Speed.} In Fig.~\ref{fig:memory_speed}, we also observe a notable inference acceleration when pruning experts from $r=6$ to $r=4$, a phenomenon not attributable to decreased inter-GPU communication overhead. We think that this enhancement stems primarily from improved temporal and spatial locality. This includes reduced cache misses, optimized memory prefetching, and faster block loading. Readers may refer to papers related to memory-intensive LLM inference (such as~\cite{2312.11514}) to find more explanation.

\textbf{More Ablations.} We carry out efficient layer-wise expert pruning in this work. Readers may be curious about the effectiveness compared with a layer-by-layer progressive searching paradigm, where the pruning of subsequent layers is aware of the pruning result of previous layers. To this end, we compare these two pruning paradigms in Tab.~\ref{tab:ablation}, leveraging the Mixtral 8x7B model and calibration samples from the C4 dataset. Although the layer-by-layer progressive manner can get slightly better results with fewer experts pruned, when it comes to a high expert pruning rate (e.g., 50\% pruning rate with $r=4$), more performance drop is observed. We attribute this to the possible overfitting of the small calibration dataset.

\section{Conclusion and Discussion}\label{conclusion}

In this paper, based on the structural characteristics of MoE LLMs and the shortcomings of current weight pruning schemes, we focus on expert-level model sparsification and, for the first time, provide post-training expert pruning together with dynamic (expert) skipping methods to enhance the deployment efficiency of MoE LLMs. Our methods can significantly reduce memory usage and enhance inference speed while maintaining high model performance. Looking ahead, we aim to further refine our pruning/skipping techniques and incorporate them with weight pruning or parameter quantization strategies, achieving more effective deploying approaches for MoE LLMs.

\section*{Limitations}

Our method can reduce memory usage and improve inference speed for more efficient deployment of MoE LLMs. Despite its advancements, there are still some limitations. Firstly, our method for expert pruning is based on the enumeration of expert combinations. This is feasible for pruning currently popular MoE LLMs with 4 or 8 experts. However, with the number of experts in each MoE layer increasing (e.g., 32 experts in one MoE layer), it will be cumbersome to perform our pruning algorithm. Secondly, we conduct experiments on the open-sourced Mixtral 8x7B and Mixtral 8x7B Instruct models as they are by far the most popular MoE LLMs. With the development of MoE LLMs, we will carry out experiments on other MoE LLMs in the future to give a more comprehensive analysis of the generalizability and scalability of our method.

\section*{Ethics Statement}

Our research focuses on improving the deployment efficiency of Mixture-of-Experts (MoE) large language models (LLMs) through expert-level sparsification techniques, aiming to reduce model sizes and enhance inference speed without compromising performance. While our methods offer potential benefits for deploying advanced LLMs more broadly and efficiently, we acknowledge the importance of considering the ethical implications of deploying such models. These include ensuring the responsible use of LLMs, mitigating biases in model outputs, and addressing privacy concerns. We commit to making our code available for transparency and encourage the community to use our findings responsibly, considering the societal impacts of deploying LLMs.

\section*{Acknowledgement}
This project is funded in part by National Key R\&D Program of China Project 2022ZD0161100, by the Centre for Perceptual and Interactive Intelligence (CPII) Ltd under the Innovation and Technology Commission (ITC)’s InnoHK, by General Research Fund of Hong Kong RGC Project 14204021. Hongsheng Li is a PI of CPII under the InnoHK. 

\bibliography{custom}
\clearpage
\appendix
\section{Appendix}
\label{sec:appendix}

\subsection{Expert Selection Tendency in MoE Models}

\begin{figure}[htbp]
    \centering
    \includegraphics[width=1.02\linewidth]{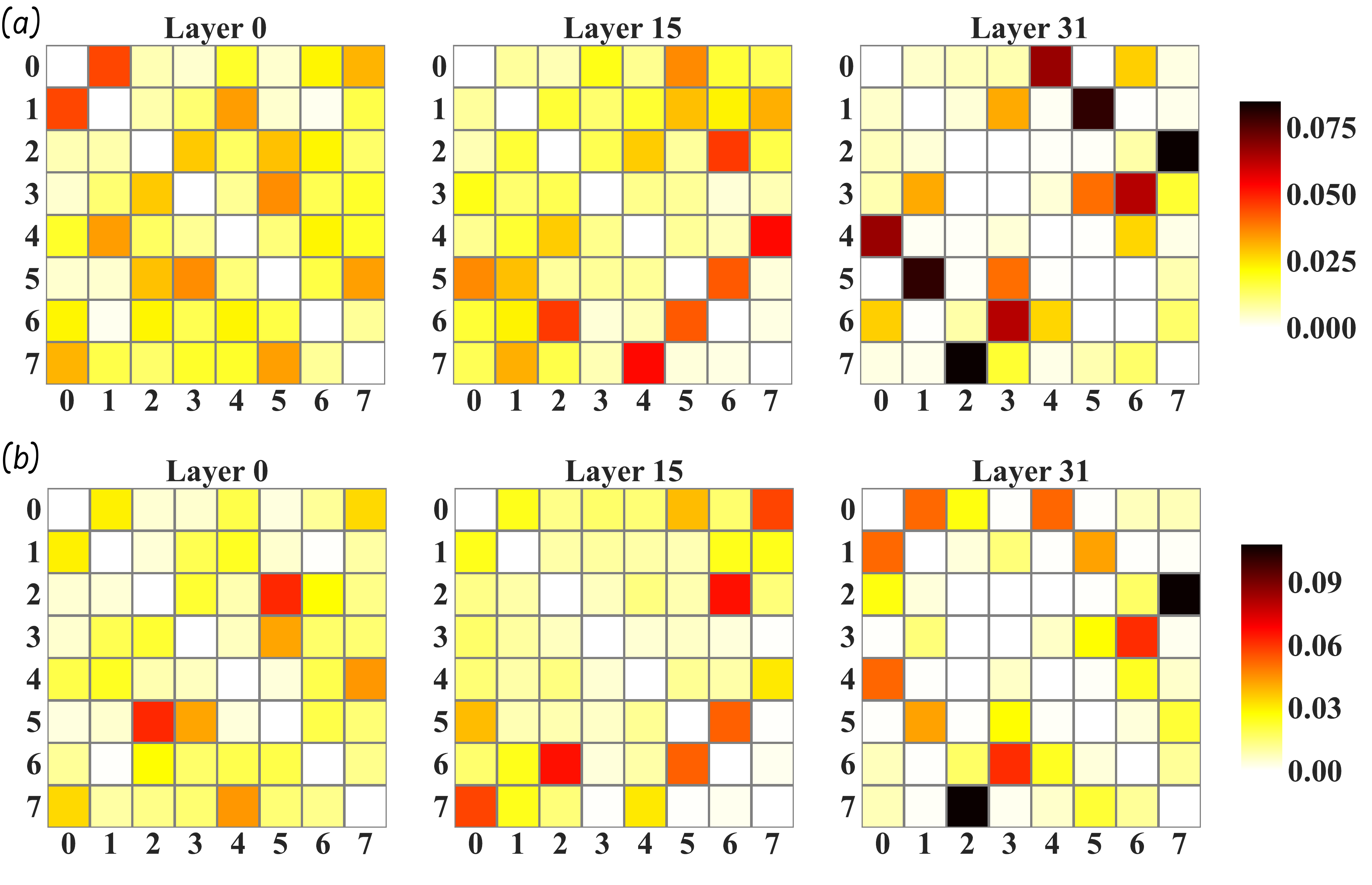}
    \vspace{-1em}
    \caption{Frequency visualization of expert selection in layer 0, layer 15, and layer 31 for the Mixtral 8x7B model on samples of (a) C4~\cite{2019t5} and (b) MATH~\cite{hendrycksmath2021} dataset respectively. The model exhibits certain preferences in the selection of experts.}
    \label{fig:expert_stat}
    \vspace{-1em}
\end{figure}

To investigate the tendency of expert selection during model inference, we analyze the Mixtral 8x7B model using samples from the C4 dataset~\cite{2019t5} (Fig.~\ref{fig:expert_stat} (a)) and the MATH dataset~\cite{hendrycksmath2021} (Fig.~\ref{fig:expert_stat} (b)), respectively. C4 is a pre-training dataset representing the general relationships between experts, MATH is a dataset designed for a specific downstream task. We visualize the probability of different experts being selected in layer 0, layer 15, and layer 31 of the model during inference. As top-2 experts are chosen by default, each grid in the plot represents the frequency of two experts being selected simultaneously during the forward pass over the sample set (the x-axis and the y-axis of the plot each represent one expert). As can be seen from Fig.~\ref{fig:expert_stat}, the model exhibits a certain tendency in the selection of experts, particularly when tailored to specific downstream tasks.

\subsection{Theoretical Insight and Broader Application of Dynamic Skipping}
\label{sec:theory_dynamic_skipping}
Consider a single MoE layer. Suppose in a top-$k$ setting with a total of $n$ experts, for every single token $\boldsymbol x$, the routing weights of the $n$ experts are $w_1, \dots, w_n, \sum_{i=1}^nw_i = 1$. Without loss of generality, we assume $w_1\ge w_2\ge\dots \ge w_k \ge \dots \ge w_n$, and the output feature vector of each expert are $\mathbf f_1, \dots, \mathbf f_k, \dots, \mathbf f_n$, where $\mathbf f_i = \mathcal{E}_{i}(\boldsymbol{x})$.

Without dynamic skipping, the original output of this layer should be:
\begin{align*}
\boldsymbol{z} = \frac{1}{\sum_{m=1}^k w_m}\sum_{m=1}^k w_m\mathbf f_m. 
\end{align*}
Suppose after dynamic skipping, only top-$i$ experts remain $(1\le i \le k)$. Then the output is:
\begin{align*}
\hat{\boldsymbol{z}} = \frac{1}{\sum_{m=1}^i w_m}\sum_{m=1}^i w_m\mathbf f_m.
\end{align*}
As defined above in the main paper, the reconstruction loss can be calculated by: 
\begin{align*}
\mathcal L = ||\hat{\boldsymbol{z}} - \boldsymbol{z}||_2.
\end{align*}
Through experiment, we observe that the distribution of $||\mathbf f_1 - \mathbf f_2||_2$ is fairly concentrated for Mixtral 8x7b on the C4 calibration set. Thus for simplification, we assume $||\mathbf f_m - \mathbf f_n||_2 (m\ne n)$ to be a fixed value, say $D$. Therefore,
\begin{equation}
\resizebox{1.02\hsize}{!}{$\begin{aligned}
\mathcal L &=||\hat{\boldsymbol{z}} - \boldsymbol{ z}||_2 \\
&= ||\frac{1}{\sum_{m=1}^i w_m}\sum_{m=1}^i w_m\mathbf f_m - \frac{1}{\sum_{m=1}^k w_m}\sum_{m=1}^k w_m\mathbf f_m||_2 \\
&= \frac{||\sum_{n=1}^k\sum_{m=1}^i w_nw_m\mathbf f_m - \sum_{n=1}^i\sum_{m=1}^k w_nw_m\mathbf f_m||_2}{(\sum_{m=1}^i w_m)(\sum_{m=1}^k w_m)} \\
&= \frac{||\sum_{n=1}^k\sum_{m=1}^i w_nw_m\mathbf f_m - \sum_{n=1}^k\sum_{m=1}^i w_nw_m\mathbf f_n||_2}{(\sum_{m=1}^i w_m)(\sum_{m=1}^k w_m)} \\
&= \frac{||\sum_{n=i+1}^k\sum_{m=1}^i w_nw_m(\mathbf f_m-\mathbf f_n)||_2}{(\sum_{m=1}^i w_m)(\sum_{m=1}^k w_m)} \\
&\le \frac{||\sum_{n=i+1}^k\sum_{m=1}^i w_nw_m||_2\cdot D}{(\sum_{m=1}^i w_m)(\sum_{m=1}^k w_m)} \\
&\quad (=\text{ holds if all }(\mathbf f_m-\mathbf f_n) \text{ are of the same direction}) \\
&=\frac{\sum_{m=i+1}^k w_m}{\sum_{m=1}^k w_m}D.\nonumber
\end{aligned}$}
\end{equation}
We set an upper bound $H\quad (H \le D)$ on $\mathcal L$ (make $\mathcal L \le H$) to trade-off between accuracy and inference speed, then have
$$\frac{\sum_{m=i+1}^k w_m}{\sum_{m=1}^k w_m}D \le H.$$
Let $\beta = \frac{H}{D}$, we have:
\begin{align*}
\sum_{m=i+1}^k w_m \le \beta\cdot \sum_{m=1}^k w_m.
\end{align*}
Therefore, in this generalized setting, dynamic skipping should reserve the top-$i^*$ experts where
\begin{align*}
&i^* = \min i\\
&\text{s.t.} \sum_{m=i+1}^k w_m \le \beta\cdot \sum_{m=1}^k w_m.
\end{align*}
Specifically, in the top-$2$ setting, for computational simplicity (reduce the use of additions), let $\beta = \frac{H}{D-H}$ (a little abuse of notation), then the dynamic skipping criteria for top-$2$ is
$$w_2 \le \beta w_1.$$

\subsection{Experiments on the Sizes of Calibration Datasets}\label{sec:calib_size}

We prune the Mixtral 8x7b model with different sizes of calibration datasets. To be specific, we randomly sample 1, 2, 4, 16, 64, and 256 sequences (each composed of 2048 tokens) from the C4 dataset to form calibration datasets. Then the model is pruned to $r=6$ and evaluated on various benchmarks. The average LM-eval results are reported in Tab.~\ref{table:calib_size}.
\begin{table}[h]\huge
\centering
\resizebox{0.55\columnwidth}{!}{
\begin{tabular}{cc}
\midrule
\textbf{Number of Sequence } & \textbf{LM-eval} \\ \midrule
\textbf{1}                     & 62.63            \\
\textbf{2}                     & 63.93            \\
\textbf{4}                     & 63.53            \\
\textbf{16}                    & 63.59            \\ \midrule
\textbf{64}                             & 64.32            \\
\textbf{128}                            & 64.22            \\
\textbf{256}                            & 63.94            \\ \midrule
\end{tabular}
}
\caption{Performances of expert pruning with different sizes of calibration datasets.}
\label{table:calib_size}
\end{table}

As can be seen, using 64 and 128 sequences can result in the highest overall results ($\ge 64$). Using a small set of sequences will possibly lead to performance degradation (especially for using just 1 sequence), but our method is somewhat robust to the size of datasets, as seen from the table.

\subsection{Dynamic Skipping for Domain-specific Tasks}\label{sec:skip_domain}

\begin{table}[t]\huge
\centering
\resizebox{\columnwidth}{!}{

\begin{tabular}{cccccc}
\midrule
\textbf{Model} & $r$ & \textbf{Pruning} & \textbf{Skipping} & \textbf{GSM8K (5-shot)} & \textbf{Speedup} \\\midrule
\multirow{5}{*}{\textbf{Mixtral 8x7B}}
& 8  & & & 58.61  & 1.00$\times$ \\\cmidrule{2-6}
& 8  & & \checkmark & 54.28  & 1.08$\times$ \\
& 6  & \checkmark & & 51.25  & 1.20$\times$ \\
& 6  & \checkmark & \checkmark & 47.16  & 1.21$\times$ \\
& 4  & \checkmark & & 37.07  & 1.29$\times$ \\
& 4  & \checkmark & \checkmark & 34.80 & 1.30$\times$ \\\midrule
\multirow{5}{*}{\textbf{\makecell{Mixtral 8x7B\\Instruct}}} & 8  & & & 63.46  & 1.00$\times$ \\\cmidrule{2-6}
& 8  & & \checkmark & 61.94  & 1.05$\times$ \\
& 6  & \checkmark & & 58.38  & 1.20$\times$ \\
& 6  & \checkmark & \checkmark & 53.98  & 1.28$\times$ \\
& 4  & \checkmark & & 47.01  & 1.28$\times$ \\
& 4  & \checkmark & \checkmark & 40.33  & 1.33$\times$ \\ \midrule  
\end{tabular}
}
\caption{Evaluation results of combining expert pruning with dynamic skipping for domain-specific tasks. Combining two expert-level sparsification methods will lead to more efficient deployment.}
\vspace{-0.6em}
\label{table:online_math}
\end{table}

We also perform dynamic expert skipping on domain-specific tasks (mathematical reasoning tasks). We calibrate $\beta$ for each layer using samples from the training set of MATH and evaluate 5-shot accuracy on the GSM8K dataset. We also test and report the token generation speed of each MoE model. The results are shown in Tab.~\ref{table:online_math}. In this case, dynamic expert skipping leads to more performance drops. But for the Mixtral 8x7B Instruct model, expert pruning with 2 experts and combining dynamic skipping also leads to the same inference speedup with pruning 4 experts, while achieving higher evaluation accuracy.

\subsection{Actual Memory Reduction}
A more detailed statistical comparison between our expert pruning method with baseline methods on the Mixtral 8x7B model is shown in Tab.~\ref{table:mem}.
\begin{table}[h]\huge
\centering
\resizebox{0.7\columnwidth}{!}{
\begin{tabular}{ccc}
    \midrule
    \textbf{Method} & \textbf{Sparsity} & \textbf{Memory (MB)} \\ \midrule
    \textbf{None}   & None ($r=8$)        & 89,926 ($100\%$)        \\ \midrule
    \textbf{Wanda}  & 2:4               & 51,214 ($57\%$)         \\ \midrule
    \textbf{Ours}   & $r=6$              & 68,383 ($76\%$)         \\ \cmidrule{2-3}
    \textbf{Ours}   & $r=4$               & 46,879 ($52\%$)         \\ \midrule
\end{tabular}
}
\caption{Memory reduction comparison of our expert pruning method with baselines on Mixtral 8x7B.}
\vspace{-0.6em}
\label{table:mem}
\end{table}

\subsection{Relationships with Other Network Pruning and Parameter Quantization Methods}

As plug-and-play techniques, both our proposed expert pruning and dynamic skipping methods are orthogonal to other model light-weighting schemes (e.g., weight pruning~\cite{frantar2023sparsegpt,sun2023simple}, token pruning~\cite{kim2022learned,ding2023prune}) and are compatible with weight quantization approaches~\cite{frantar2022gptq,lin2023awq}. 

\subsection{More Experiment Details}

In this part, we give more experimental details for a better understanding of our proposed methods.

\textbf{Calibration Set Construction for Expert Pruning. }For task-agnostic models, we use the samples from C4~\cite{2019t5} as the calibration dataset. Following the setting of Wanda, we sample from the first part of the training data\footnote{\url{https://huggingface.co/datasets/allenai/c4/blob/main/en/c4-train.00000-of-01024.json.gz}}. For task-specific (mathematics) models, we use samples from the training set of MATH~\cite{hendrycksmath2021}. The structure of the MATH dataset is different from C4, so we reconstruct the dataset in the format of C4 and randomly sample from it.

\textbf{Calibration Set Construction for Dynamic (Expert) Skipping. }To calculate $\beta$ for dynamic expert skipping in each MoE layer, we forward the MoE model over the calibration dataset and set $\beta$ as the median value of $\frac{w_{e_1}}{w_{e_0}}$ separately for each layer. We choose to use the median value over the calibration dataset as in this case, the skipping will happen with around 50\% possibility. Here we provide the value of $\beta$ for the Mixtral 8x7B model with calibration data sampled from C4 and MATH respectively. As can be seen, the parameter in each layer differs significantly.

C4: 0.402,0.494,0.463,0.484,0.478,0.491,0.523,\\0.521,0.544,0.570,0.574,0.489,0.503,0.618,0.568,\\0.535,0.559,0.519,0.537,0.487,0.469,0.461,0.461,\\0.469,0.458,0.418,0.433,0.418,0.406,0.433,0.447,\\0.535

MATH: 0.503,0.586,0.505,0.531,0.509,0.422,\\0.511,0.461,0.447,0.478,0.529,0.454,0.472,0.531,\\0.499,0.486,0.503,0.491,0.430,0.440,0.402,0.423,\\0.386,0.407,0.395,0.354,0.340,0.351,0.334,0.368,\\0.365,0.346 

\textbf{Model Fine-tuning. }In the part of task-specific expert pruning for domain-specific tasks, we fine-tune the Mixtral 8x7B and Mixtral 8x7B Instruct models with 8 experts, 7 experts, and 6 experts on the MetaMathQA~\cite{yu2023metamath} dataset. The training is conducted on 16 A100-80G GPUs. We train the model for 900 steps, using a learning rate of 2e-5 with the cosine learning rate scheduler.

\end{document}